\documentclass{article}

\usepackage[preprint, nonatbib]{neurips_2019_ml4ps}

\usepackage[utf8]{inputenc} 
\usepackage[T1]{fontenc}    
\usepackage{hyperref}       
\usepackage{url}            
\usepackage{booktabs}       
\usepackage{amsfonts}       
\usepackage{nicefrac}       
\usepackage{microtype}      

\usepackage{graphicx}
\usepackage{xspace}
\usepackage{amsmath}
\usepackage{tcolorbox}
\usepackage{floatrow}
\usepackage{comment}
\usepackage{wrapfig}
\usepackage{tabularx}
\usepackage{textgreek}

\title{JANUS: Parallel Tempered Genetic Algorithm Guided by Deep Neural Networks for Inverse Molecular Design}

\author{%
  {\large AkshatKumar Nigam$^{1,2,3,\dagger}$, Robert Pollice$^{2,3,\dagger}$}\\
  {\textbf{Al\'{a}n Aspuru-Guzik}$^{2,3,4,5,*}$} \\
  \\

  $^1$Department of Computer Science, Stanford University, USA.\\  
  $^2$Department of Chemistry, University of Toronto, Canada.\\
  $^2$Department of Computer Science, University of Toronto, Canada.\\
  $^4$Vector Institute for Artificial Intelligence. \\
  $^5$Lebovic Fellow, Canadian Institute for Advanced Research (CIFAR). \\
  $^{\dagger}$These authors contributed equally \\
  $^*$Correspondence to: aspuru@utoronto.ca \\
  Code available at: \href{https://github.com/aspuru-guzik-group/JANUS}{\color{blue}{https://github.com/aspuru-guzik-group/JANUS}}

}

\begin{document}

\maketitle

\begin{abstract}
Inverse molecular design, i.e., designing molecules with specific target properties, can be posed as an optimization problem. High-dimensional optimization tasks in the natural sciences are commonly tackled via population-based metaheuristic optimization algorithms such as evolutionary algorithms. However, expensive property evaluation, which is often required, can limit the widespread use of such approaches as the associated cost can become prohibitive. Herein, we present JANUS, a genetic algorithm that is inspired by parallel tempering. It propagates two populations, one for exploration and another for exploitation, improving optimization by reducing expensive property evaluations. Additionally, JANUS is augmented by a deep neural network that approximates molecular properties via active learning for enhanced sampling of the chemical space. Our method uses the SELFIES molecular representation and the STONED algorithm for the efficient generation of structures, and outperforms other generative models in common inverse molecular design tasks achieving state-of-the-art performance. \end{abstract}

\section{Introduction}
Molecular design workflows usually consist of design-make-test-analyze cycles \cite{wesolowski2016strategies, schneider2020rethinking}. Classically, the first step requires input from human scientists to propose molecules that are to be made and tested. The field of cheminformatics generated powerful computer workflows for the generation of molecular structures with desired properties mimicking human designers. In recent years, the advent of machine learning (ML) in chemistry has reinvigorated these efforts and sparked the development of numerous new data-driven tools for inverse molecular design \cite{sanchez2018inverse, pollice2021data, nigam2021assigning}.

In this work, we present JANUS, a genetic algorithm (GA) based on the SELFIES representation of molecules \cite{krenn2020self, nigam2019augmenting}. Our algorithm makes use of STONED \cite{nigam2021beyond} for the efficient generation of new molecules, obviating the need for explicit structural validity checks. Inspired by parallel tempering \cite{earl2005parallel, hansmann1997parallel}, JANUS maintains two populations of molecules that can exchange members and have distinct sets of genetic operators, one explorative and the other exploitative. The exploitative genetic operators use molecular similarity as additional selection pressure. The explorative genetic operators are guided by selection pressure from a deep neural network (DNN), which is trained by active learning over all previous generations. While our algorithm does not require domain knowledge, we implemented the option to extract structure derivation rules automatically from reference structures and bias the molecular generation. JANUS outperforms other generative models in various benchmarks such as the maximization of the penalized logarithm of the octanol-water partition coefficient scores (penalized log P) \cite{gomez2018automatic} and the minimization of molecular docking scores for various protein targets \cite{cieplinski2020we}.

\section{Related Works}
Inverse molecular design creates structures with desired properties by inverting the classical design workflow that relies on structure first and property second. The idea is to specify the target properties and explore the chemical space systematically to find structures that optimize them. Various approaches have been applied for inverse molecular design tasks including variational autoencoders (VAEs) \cite{kingma2013auto}, generative adversarial networks (GANs) \cite{goodfellow2014generative}, autoregressive models (AMs) \cite{hochreiter1997long}, flow-based generative models (FGMs) \cite{kingma2018glow}, reinforcement learning (RL) \cite{sutton2018reinforcement}, sampling of Markov decision processes (MDPs) \cite{howard1960dynamic} and metaheuristic optimization algorithms \cite{metag, metag_1}. The following sections summarize recent works in the field of inverse molecular design, and the corresponding models are used as baselines for benchmarking.

\textbf{Deep Generative Models: }\\ 
VAEs, GANs, AMs, MDPs and RL approaches belong to the group of deep generative models as they all rely on DNNs for molecular generation. VAEs convert discrete representations into continuous latent spaces and vice versa. VAEs are jointly trained with property predictors for reliably reconstructing the provided input, and arranging the latent space based on molecular properties. This allows for both gradient-based optimization in the latent space and sampling the vicinity of regions with desired property values to generate molecules with favorable properties. Since their first demonstration in chemistry \cite{gomez2018automatic}, they have received widespread attention, notably being applied to the design of nanoporous crystalline reticular materials \cite{yao2021inverse}, the optimization of binding affinities for drug discovery \cite{boitreaud2020optimol}, the exploration of inorganic materials \cite{pathak2020deep} and scaffold-based drug design \cite{lim2020scaffold}. Common implementations of VAEs for inverse molecular design include CVAE \cite{gomez2018automatic}, GVAE \cite{kusner2017grammar} and SD-VAE \cite{dai2018syntaxdirected}, which operate on molecular string representations. Alternatively, JT-VAE  \cite{jin2018junction}, CGVAE \cite{liu2018constrained} and DEFactor \cite{assouel2018defactor} use matrix representations of molecular graphs. DEFactor is particularly notable as it uses a differentiable model architecture. Furthermore, the framework of VAEs can be generalized to encoder-decoder architectures where the output of the decoder does not correspond to the input of the encoder. This has been used in translation tasks and can be used to propose structural modifications, and VJTNN \cite{jin2018learning} follows this approach by implementing a graph-to-graph translation.

GANs are characterized by joint adversarial training of a generator and a discriminator DNN. The generator DNN proposes molecular structures from a high-dimensional latent space, while the discriminator distinguishes the proposed structures as either originating from a reference dataset or as newly generated. The generator and discriminator are trained as competing networks. For inverse design, ORGAN \cite{guimaraes2017objective} and ORGANIC \cite{sanchez2017optimizing} were the first implementations of a GAN for molecular design that were trained on molecular string representations with RL. Follow up work includes the use of adjacency matrices \cite{de2018molgan, maziarka2020mol}. AMs learn to predict future states based on previous outputs and are typically used for sequences. Recurrent neural networks (RNNs) \cite{mandic2001recurrent} are the most commonly used examples for the sequential generation of string-based molecular graph representations \cite{segler2018generating}. MRNN \cite{popova2019molecularrnn} combines an RNN for molecule generation and uses it as a policy network in an RL framework for the optimization of molecular properties. RL approaches in general rely on learning how to take actions in an environment to maximize a cumulative reward. Accordingly, the goal is to use DNNs to predict the best action based on a given state. GCPN \cite{you2018graph} and PGFS \cite{gottipati2020learning} use matrix representations of molecular graphs, while MolDQN \cite{zhou2019optimization} and REINVENT \cite{olivecrona2017molecular} use string-based representations of molecular graphs. GCPN relies on an MDP for generating new molecules from initial ones. MolDQN utilizes an action space of elementary modifications chosen by domain experts. Similarly, PGFS also implements an expert-derived action space but in the form of reaction templates to mimic forward synthesis for molecular generation. In contrast, REINVENT uses an RNN framework to propose new molecules.

In contrast to AMs, MDPs require explicit actions from a decision maker and knowledge of the current state to predict future states. MDPs are characterized by the dependence of the next state of the system only on the current state and the action of the decision maker, and are hence independent of all previous states and actions. Monte Carlo tree search (MCTS) is a method to sample decision processes such as MDPs that has gained significant popularity in ML due to its use in AlphaGo in combination with DNNs \cite{silver2016mastering}. This also inspired its application for inverse molecular design. Particular implementations include ChemTS \cite{yang2017chemts} that relies on string-based representations of molecular graphs and an RNN for modeling the MCTS. Alternative approaches are unitMCTS \cite{rajasekar2020goal} and MARS \cite{xie2021mars}, which both rely on matrix representations of molecular graphs. While unitMCTS implements MCTS for modeling the molecular generation process, MARS uses annealed Markov chain Monte Carlo sampling. Finally, FGMs are based on normalized flow to model probability distributions by using a sequence of invertible functions acting on an input to explicitly model the log-likelihood. They can be used as an alternative to VAEs to generate continuous latent spaces with the advantage that only one direction of the transformation needs to be trained due to its inherent invertibility. One model relying on FGMs is GraphAF \cite{shi2020graphaf}, which combines the advantages of FGMs with the ones of AMs and uses an RL framework for optimization of molecular properties.

\textbf{Metaheuristic Optimization Algorithms:}\\
GAs belong to the large family of population-based metaheuristic optimization algorithms. They have a long-standing track record in the natural sciences for tackling complicated design problems \cite{holland1992genetic}. Accordingly, GAs have been applied to molecular design tasks for decades, and they remain one of the most popular metaheuristic optimization algorithms for that purpose. The main difference between previously published approaches lies in the specific implementation of the genetic operators, particularly the mutation and crossover of molecules. Most prevalent is the use of expert rules to ensure the validity of the generated structures. Particular examples of GAs implementing expert molecular generation rules include the GB-GA \cite{jensen2019graph}, CReM \cite{polishchuk2020crem}, EvoMol \cite{leguy2020evomol} and Molfinder \cite{kwon2021molfinder}. However, recently, the robustness of SELFIES has been exploited in the GA-D \cite{nigam2019augmenting} model to be able to rely only on random string modifications for mutations obviating the definition of hand-crafted structure modification rules. An alternative approach that also belongs to the family of metaheuristic optimization algorithms is MSO \cite{winter2019efficient}, which relies on particle swarm optimization over the continuous latent space derived from a VAE. Furthermore, mmpdb \cite{dalke2018mmpdb} is an implementation of matched molecular pair analysis (MMPA \cite{kenny2005structure}) and relies on structure modification rules explicitly derived from a dataset of molecule pairs. As the derived rules are applied in a stochastic way the use of mmpdb for molecular design can be considered a metaheuristic optimization algorithm.

\section{Architecture}
\subsection{Overview}
\begin{figure}[htbp!]
\centering
\includegraphics[width=1.00\textwidth]{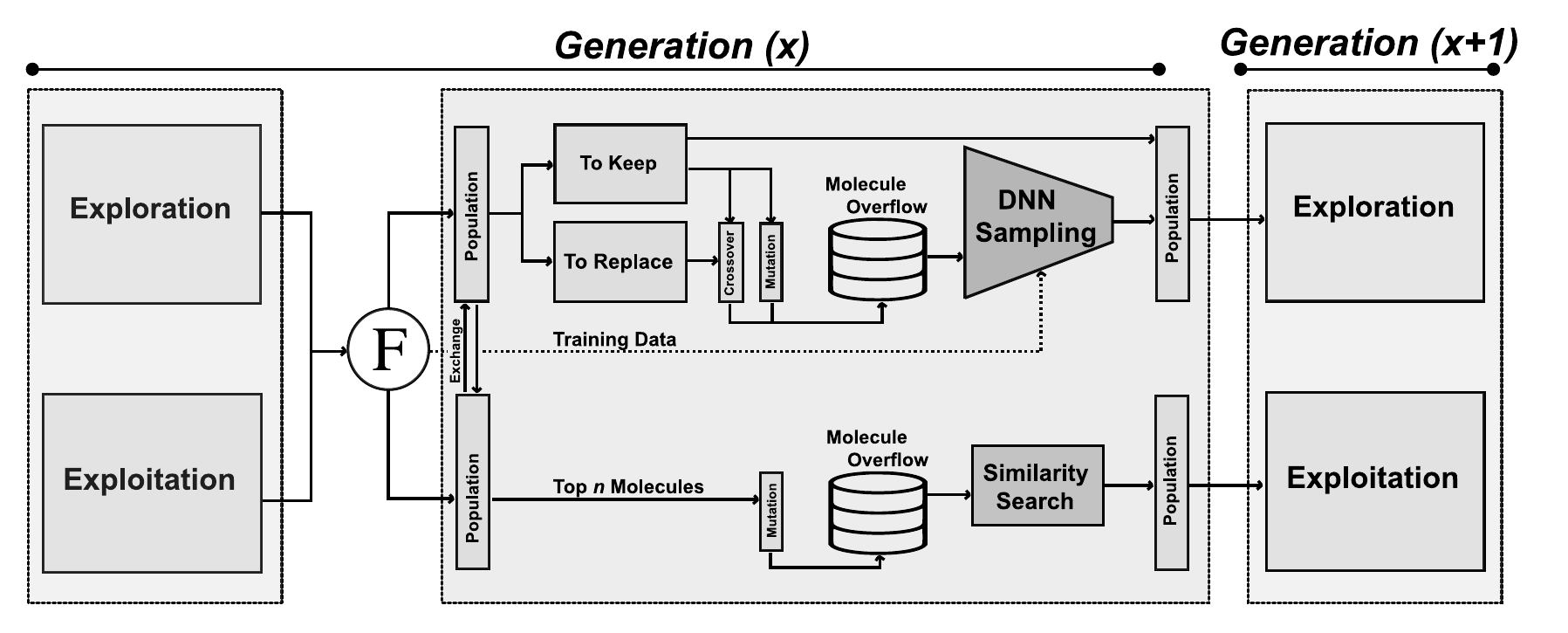}
\caption{Schematic depiction of the architecture of JANUS. Two populations are propagated in parallel with distinct sets of genetic operators. The exploitative population uses molecular similarity as additional selection pressure, the explorative population uses a deep neural network (DNN) estimating molecular properties as additional selection pressure.\label{fig:janus_scheme}}
\end{figure}

JANUS is a genetic algorithm that is initiated either with random structures or with a provided set of molecules. At every iteration or generation, two molecule populations of fixed size are maintained. Members in each population compete against each other to survive and proceed to the next generation. The quality of a molecule is measured by the fitness function. Between consecutive generations, JANUS uses elitist selection \cite{baluja1995removing}, i.e., molecules with higher fitness remain undisturbed, while structures with low fitness are replaced with new members obtained via genetic operators. The sets of genetic operators differ between the two populations, one being explorative and the other exploitative. In addition, an overflow of potential children is produced from the parent molecules that are filtered based on additional selection pressure. Only the most suitable children are used in the subsequent generation. In the following sections, we will describe the most important components of JANUS (cf. Figure \ref{fig:janus_scheme}) in more detail.

\subsection{Genetic Operators}
The implementation of mutations and crossovers in GAs often requires expert design. In contrast, JANUS relies on the STONED algorithm \cite{nigam2021beyond} for efficient molecule generation. Namely, as demonstrated in Figure \ref{fig:stoned_scheme}(a), for mutating a given molecule, random modifications (by way of character deletions, additions, or replacements) are performed on the SELFIES representing the molecule. To increase the diversity of the mutated structures, we follow the suggestion of Nigam et al. \cite{nigam2021beyond} and use multiple reordered SMILES representations of the same molecule to generate numerous alternative SELFIES.

Starting from two molecules, an apt crossover molecule resembles both parent molecules. Therefore, as implemented in STONED, we form multiple paths between the two molecules (again obtained via random string modifications to the SELFIES representations, as demonstrated in Figure \ref{fig:stoned_scheme}(b)) and select structures that possess high joint molecular similarity. For the parent molecules $M$ = $\{m_1, m_2, ...\}$, a good crossover molecule ($m$) maximizes the following joint similarity function (cf. Nigam et al. \cite{nigam2021beyond}):

\begingroup\makeatletter\def\f@size{10.0}\check@mathfonts
\def\maketag@@@#1{\hbox{\m@th\large\normalfont#1}}%
\begin{align}
F(m) = \frac{1}{n} \sum\limits_{i=1}^n\text{sim}(m_i, m)  - \left[\max_{i}(\text{sim}(m_i, m)) - \min_{i}(\text{sim}(m_i, m))\right]
\end{align}\endgroup

Consideration of multiple SELFIES representations, multiple string modifications and the formation of numerous paths leads to the generation of many children molecules during a single genetic operation via mutation or crossover.

\subsection{Selection Pressure}
To deal with the overflow of molecules generated by the genetic operators, we propose a strategy to add additional selection pressures to the population when selecting children to propagate to the next generation. At the end of every generation, we train a DNN on molecules with known fitness values. We explore both training a neural network for accurately predicting the fitness function (abbreviated as P herein) and for separating the good molecules (i.e., high fitness) from the bad (i.e., low fitness) via a classifier (abbreviated as C herein). The trained model is used to evaluate the overflow molecules for the explorative population, and the most promising structures (i.e., possessing the highest estimated values) are added to the main population. In our experiments, we compare the use of a DNN for applying selection pressure to the strategy of randomly sampling children from the overflow molecules. For the exploitative population, additional selection pressure is added by only propagating the molecules most similar to the parents onto the next generation. Similarity is assessed via the Tanimoto coefficient based on circular molecular fingerprints \cite{rogers2010extended}.

\subsection{Parallel Populations}
In every generation, two populations of molecules are maintained. While one explores the chemical space, the other exploits the regions that possess high fitness values. New molecules in the exploration population are obtained using both the mutation and crossover operations. Selection pressure is then applied using a DNN upon the overflow molecules, leading to a new population in the next generation. For the exploitation population, novel molecules are obtained using solely the mutation operation. Subsequently, selection pressure is applied by picking molecules with high similarity to the parents. Additionally, within each generation, the two populations exchange members based on their fitness. Molecules of the explorative population with high fitness enter the exploitative population to investigate the corresponding regions of chemical space more extensively. In turn, molecules of the exploitative population with high fitness enter the explorative population to explore regions of chemical space further away from current well-performing molecules. Additionally, the molecules that are selected from the exploration population for proceeding to the genetic operators are not solely chosen based on fitness but also based on a parameter $F_{25}$ that essentially determines the temperature of the population. At higher temperature, molecules of lower fitness are more likely to proceed to the next generation. Inspired by Fermi-Dirac statistics \cite{dirac1926theory, fm_126}, we evaluate the relative frequency $p_i$ for molecule $i$ to be selected according to the following formula:

\begin{equation}
p_i = \left(3^{\frac{F_{50} - F_i}{F_{50} - F_{25}}} + 1 \right)^{-1}\;
\label{eq:fermi_dirac_selection}
\end{equation}

In Equation \ref{eq:fermi_dirac_selection}, $F_{50}$ corresponds to the fitness of the n\textsuperscript{th} molecule when the population is sorted by decreasing fitness and n molecules need to be selected in total. This fitness corresponds to a relative frequency of 0.50 for being selected. The parameter $F_{25}$ is the fitness value that is to be assigned a relative frequency of 0.25 for being selected. The lower it is, the higher the effective temperature of the system and the more likely molecules of very low fitness are propagated to the next generation increasing the extent of exploration.

\section{Experiments}
In the following sections, we compare the performance of JANUS against several baseline methods in various established molecular design benchmark tasks. Notably, the property distributions of molecules generated from random SELFIES depend on the particular version of SELFIES used (cf. Figure \ref{fig:random_selfies}), which is mainly caused by extensions to the SELFIES alphabet. In all our optimizations, we use SELFIES version 1.0.3 which leads to a slight decrease in performance but provides faster encoding and decoding of molecules. In addition to the experiments presented in the main text, we also carried out the GuacaMol suite of benchmarks\cite{brown2019guacamol} without using a neural network to apply selection pressure. Importantly, JANUS achieves state-of-the-art on one half of the benchmarks and close to state-of-the-art on the other half making it overall competitive with alternative models in the literature.\cite{winter2019efficient,kwon2021molfinder,leguy2020evomol,NEURIPS2020_8ba6c657} In particular, to the best of our knowledge, JANUS shows the second best performance of any generative model not relying on neural networks for molecule generation.

\subsection{Unconstrained Penalized log P Optimization}

First, we investigated the performance of JANUS for maximizing the penalized logarithm of the octanol-water partition coefficient scores, referred to as penalized log P score J(m) \cite{gz2017automatic}:

\begin{equation}
J(m) = \text{log P}(m) - \text{SAscore}(m) - \text{RingPenalty}(m)\;
\label{eq:penalized_logp}
\end{equation}

For molecule $m$, P($m$) corresponds to the octanol-water partition coefficient, SAscore($m$) is the synthetic accessibility score \cite{ertl2009estimation}, and Ring Penalty($m$) linearly penalizes the presence of rings of size larger than 6. For comparison, we consider baselines that produce SMILES that are limited to 81 characters \cite{yang2017chemts} and normalize the log P, SAS and RingPenalty scores based on a subset of the ZINC dataset \cite{gz2017automatic}. We observe that this 81 character restriction is important for comparison, as without this limit, significantly higher penalized log P scores can be generated by having longer SMILES strings. We compare both the averages of the best scores found in several independent runs and single best scores found within one particular run as both types of scores have been reported in the literature.

\begin{figure}[htbp!]
\centering
\begin{tabular}{ ll }
(a) & (b) \\
\includegraphics[width=0.47\textwidth]{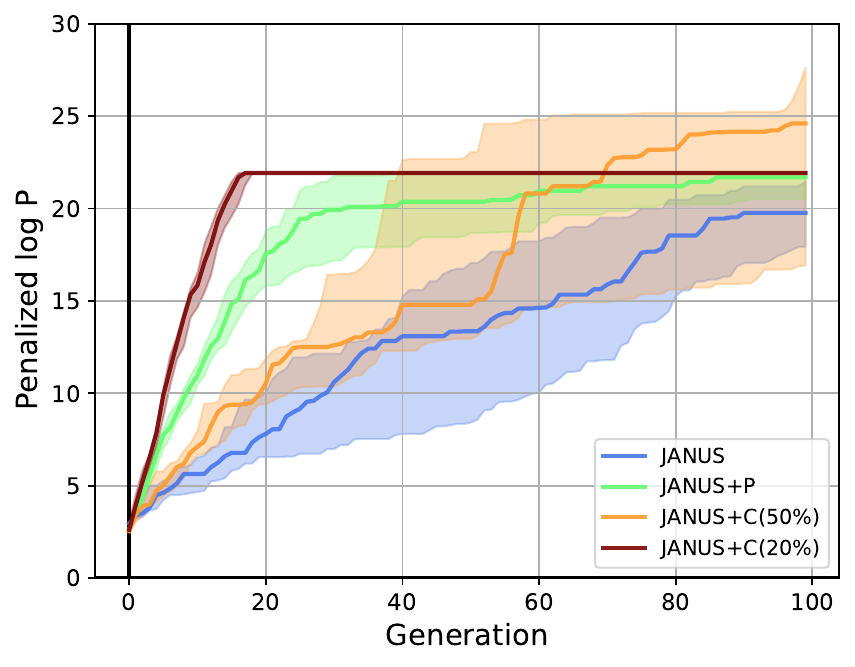} %
& %
\includegraphics[width=0.47\textwidth]{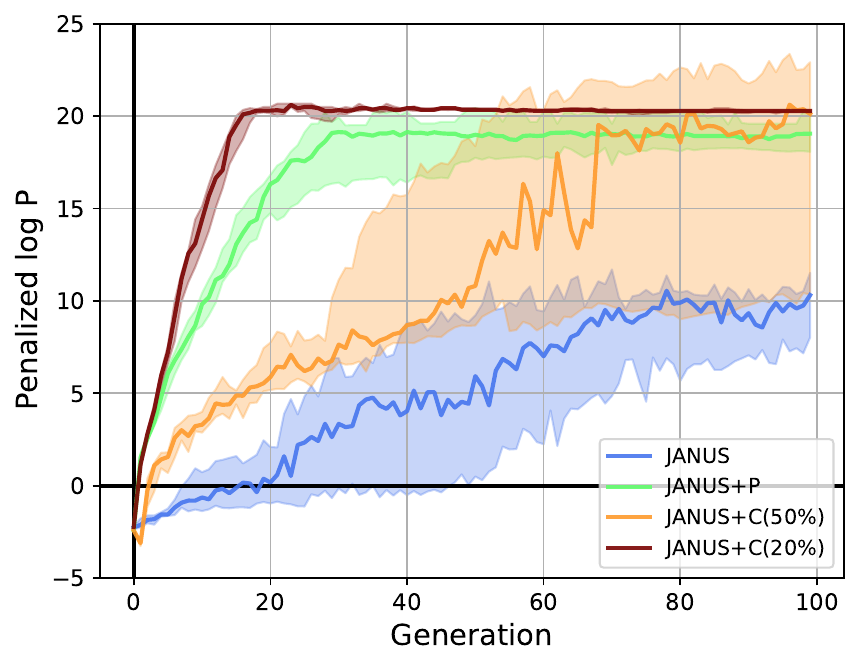}
\end{tabular}
\caption{Optimization progress of JANUS with four variations of selection pressure in the maximization of the penalized logarithm of the octanol-water partition coefficient (penalized log P). (a) Progress of the median of the highest fitness in each generation across 15 independent runs. (b) Progress of the median-of-medians fitness in each generation across 15 independent runs. The semi-transparent areas in both (a) and (b) depict the fitness intervals between the corresponding first and third quantiles.\label{fig:unconstrained_logp}}
\end{figure}

The performance of JANUS in comparison to literature baselines is summarized in Table \ref{tab:unconstrained_logp}. Notably, we run JANUS with four different variations of the selection pressure, first without any additional selection pressure (i.e., randomly sampling from the overflow) in the exploration population (“JANUS”), second with selection pressure from a DNN predictor trained to predict the fitness of molecules (“JANUS+P”), and third with selection pressure from a DNN classifier trained to distinguish between high and low fitness molecules (“JANUS+C”). Furthermore, for the classifier we also compare the performance of classifying the top 50\% of all molecules as having a high fitness (“JANUS+C(50\%)”) against only classifying the top 20\% of all molecules as having a high fitness (“JANUS+C(20\%)”). The optimization progress of JANUS with these four variations of the selection pressure is depicted in Figure \ref{fig:unconstrained_logp}.

\begin{table}[htbp!]
\caption{Comparison of JANUS against literature baselines for the maximization of the penalized logarithm of the octanol-water partition coefficient scores. The entry denoted as “JANUS” does not use additional selection pressure for the exploration population, “JANUS+P” uses a DNN predictor as additional selection pressure, the two “JANUS+C” entries use a DNN classifier as additional selection pressure.}
\resizebox{1.\textwidth}{!}{
\begin{tabular}{rcc rcc} 
\toprule
\textbf{Algorithm} & \textbf{Average of Best} & \textbf{Single Best} & \textbf{Algorithm} & \textbf{Average of Best} & \textbf{Single Best} \\
\cmidrule(lr){1-3}
\cmidrule(lr){4-6}
GVAE \cite{kusner2017grammar}  &  $2.87 \pm 0.06$ &  -  & GraphAF \cite{shi2020graphaf}  &  - &  12.23  \\
SD-VAE \cite{dai2018syntax}  &  $3.60 \pm 0.44$ &  -  &  GB-GA \cite{jensen2019graph}  &  $15.76 \pm 5.76$&  -  \\
ORGAN  \cite{guimaraes2017objective} &  $3.52 \pm 0.08$ &  -  &  GA+D(t)\small $^{\text{1}}$ \cite{nigam2019augmenting}  &  $20.72 \pm 3.14$&  -  \\
CVAE + BO  \cite{G_mez_Bombarelli_2018}  &  $4.85 \pm 0.17$ &  -  &  GEGL\small $^{\text{2}}$ \cite{NEURIPS2020_8ba6c657}  &  $31.40 \pm 0.00$&  31.40  \\
JT-VAE \cite{jin2018junction}  &  $4.90 \pm 0.33$ &  -  & \multicolumn{3}{c}{\rule{0.57\textwidth}{1pt}} \\
ChemTS \cite{yang2017chemts}  &  $5.6 \pm 0.5$ &  -  & \textbf{JANUS}\small $^{\text{3}}$  &  $\mathbf{18.4 \pm 4.4}$ &  \textbf{20.96}  \\ 
GCPN \cite{you2018graph} &  $7.87 \pm 0.07$ &  -  & \textbf{JANUS+P}\small $^{\text{3}}$  &  $\mathbf{21.0 \pm 1.3}$ &  \textbf{21.92}  \\  
MRNN \cite{popova2019molecularrnn} &  - &  8.63 & \textbf{JANUS+C(50\%)}\small $^{\text{3}}$ &  $\mathbf{23.6 \pm 6.9}$ &  \textbf{34.04}  \\  
MolDQN \cite{zhou2019optimization} &  - &  11.84  & \textbf{JANUS+C(20\%)}\small $^{\text{3}}$ &  $\mathbf{21.9 \pm 0.0}$ &  \textbf{21.92}  \\  
\bottomrule
\end{tabular}}
\begin{center}
\scriptsize $^{\text{1}}$: Average of 5 separate runs with 500 molecules of up to 81 SMILES characters per generation and 1000 generations. \\
\scriptsize $^{\text{2}}$: Average of 5 separate runs with 16,384 molecules of up to 81 SMILES characters per generation and 200 generations. \\
\scriptsize $^{\text{3}}$: Average of 15 separate runs with 500 molecules of up to 81 SMILES characters per generation and 100 generations. \\
\end{center}
\label{tab:unconstrained_logp}
\end{table}

As can be seen in Table \ref{tab:unconstrained_logp}, JANUS with all four variations of selection pressure achieves state-of-the-art performance in the maximization of the penalized log P. It outperforms all but two alternative approaches significantly in terms of both average performance and single best molecule found. The GA+D approach reaches similarly high results but after 10 times the number of generations. Only GEGL performs better on average than any of our approaches, but with double the number of generations and more than 30 times the number of structures generated per generation leading to more than 60 times the number of fitness evaluations. JANUS with additional selection pressure still finds molecules with significantly higher penalized log P. Additionally, while the average of the maximum penalized log P across independent runs does not show a very large difference after 100 generations for the variants of JANUS tested, both the optimization progress as illustrated in Figure \ref{fig:unconstrained_logp}(a) and the single best-performing molecule found shows very significant differences. In particular, using selection pressure via the DNN predictor (“JANUS+P”) or via the DNN classifier (“JANUS+C”) results in populations with higher median fitness than without additional selection pressure (“JANUS”) as demonstrated in Figure \ref{fig:unconstrained_logp}(b). Additionally, the selection pressure from the DNN predictor results in a fast increase in fitness very early on as the predictor learns the characteristics of a molecule with a high penalized log P score. The respective runs converge to the best-performing linear alkane with 81 carbon atoms and a fitness of 21.92, which corresponds to a local optimum. When selection pressure is applied via a DNN classifier, the optimization progress shows a large dependence on the fraction of molecules that are considered top-performing. When only the top 20\% are considered top-performing the respective runs converge to the best-performing linear alkane even faster than with the DNN predictor. However, as can be seen in the very narrow fitness intervals of the generations in Figure \ref{fig:unconstrained_logp}(b), the exploration of the chemical space is extremely limited leading to convergence to the local optimum easiest to find. When the top 50\% are considered top-performing the exploration is significantly enhanced resulting in a larger variation of fitness values across independent runs. This indicates that tuning hyperparameters in the DNN classifier can be used to switch between exploration and exploitation.

\begin{figure}[htbp!]
\centering
\includegraphics[width=1.00\textwidth]{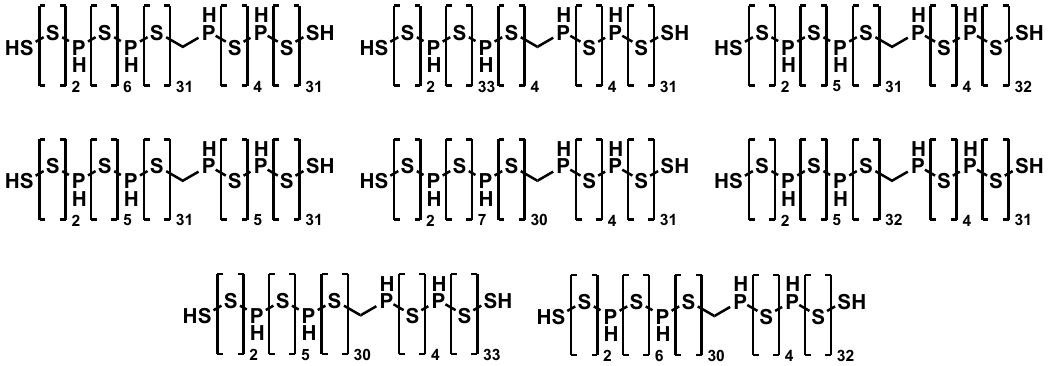}
\caption{Molecules discovered by JANUS with the highest penalized log P score of 36.62 that are within the 81 SMILES character limit.\label{fig:unconstrained_logp_molecules}}

\end{figure}

To the best of our knowledge, these runs also identified the best-performing molecules ever found by inverse molecular design algorithms that fulfill the 81 SMILES character limit, even outperforming GEGL significantly. When JANUS (specifically “JANUS+C(50\%)”) was allowed to run for more than 100 generations, a penalized log P value of 36.62 was achieved at generation 118 in one particular run. The fitness did not increase thereafter, but several isomers with the same fitness were discovered. The corresponding structures are depicted in Figure \ref{fig:unconstrained_logp_molecules}. In recent years, it has been claimed that the unconstrained penalized log P maximization task is not useful for benchmarking inverse molecular design algorithms as it has trivial solutions simply producing molecules with ever longer chains \cite{xie2021mars}. However, that is only true if no SMILES character limit is enforced. When the 81 character limit is enforced, the trade-off between the log P and the SAscore terms results in highly non-trivial solutions as evident from Figure \ref{fig:unconstrained_logp_molecules}. It is important to realize that these structures are very non-intuitive to design as the exact placement of the non-sulfur atoms in the chain can affect the total score considerably. In that regard, it is notable that JANUS even outperforms the human design demonstrated by Nigam et al. \cite{nigam2019augmenting} proposing a linear chain of sulfur atoms terminated by thiol groups with a penalized log P score of 31.79. This shows that the incorporation of non-sulfur atoms at specific positions is highly non-intuitive. Apart from this unconstrained experiment, we also ran a variant where penalized log P scores are improved for select molecules with the added constraint of keeping the similarity to the initial structure high (cf. Section \ref{sec:constrained_logp} for details).

\subsection{Imitated Inhibition}

Next, we tested the performance of JANUS for the imitated protein inhibition tasks introduced by Jin et al. \cite{jin2020multi}. In these tasks, the objective is to produce 5,000 molecules that inhibit either only GSK3{\textbeta} (A in Table \ref{tab:imitated_inhibition}), or only JNK3 (B in Table \ref{tab:imitated_inhibition}), or both (C in in Table \ref{tab:imitated_inhibition}). A fourth task is to not only generate molecules that inhibit both GSK3{\textbeta} and JNK3, but also have high drug-likeness (i.e., QED $\geq$ 0.6) \cite{bickerton2012quantifying} and high synthesizability (SAscore $\leq$ 4.0) \cite{ertl2009estimation}. Whether a molecule inhibits these targets is assessed via a random forest classifier model trained by Jin et al. \cite{jin2020multi} that imitates docking. Accordingly, values above 0.5 are classified as inhibition. The performance in these tasks is assessed via success, novelty and diversity metrics \cite{jin2020multi}. The success metric is the fraction of the molecules in the 5,000 produced that fulfill the constraints. The novelty metric is based on the similarity of the generated molecules to a reference dataset. The diversity metric is based on the similarity among the generated molecules (Details in the Supporting Information).

\begin{table}[htbp!]
\caption{Comparison of JANUS against literature baselines for the four imitated inhibition benchmark tasks of the targets GSK3{\textbeta} and JNK3 (A: GSK3{\textbeta}, B: JNK3, C: GSK3{\textbeta} + JNK3, D: GSK3{\textbeta} + JNK3 + QED + SAscore). These benchmarks are evaluated based on 5,000 molecules generated by the model. The entry denoted as “JANUS” does not use additional selection pressure for the exploration population, “JANUS+P” uses a DNN predictor as additional selection pressure, the “JANUS+C(50\%)” entry uses a DNN classifier as additional selection pressure.}
\resizebox{1.\textwidth}{!}{
\begin{tabular}{r cccc cccc cccc}
\toprule
\textbf{Method} & \multicolumn{4}{c}{\textbf{Success}} & \multicolumn{4}{c}{\textbf{Novelty}} & 
\multicolumn{4}{c}{\textbf{Diversity}} \\
\cmidrule(lr){2-5}
\cmidrule(lr){6-9}
\cmidrule(lr){10-13}
       & \textbf{A}       & \textbf{B}       & \textbf{C}     & \textbf{D}  & \textbf{A}       & \textbf{B}       & \textbf{C}     & \textbf{D}  & \textbf{A}        & \textbf{B}        & \textbf{C}  & \textbf{D}     \\
\midrule
JTVAE \cite{jin2018junction} & 32.2\% & 23.5\% & 3.3\% & 1.3\% & 11.8\% & 2.9\% & 7.9\% & - & 0.901 & 0.882 & 0.883 & -  \\ 
GCPN  \cite{you2018graph} & 42.4\%  & 32.3\% & 3.5\% & 4.0\% & 11.6\% & 4.4\% & 8.0\% & - & 0.904 & 0.884 & 0.874 & -  \\
GVAE-RL \cite{jin2020multi} & 33.2\% & 57.7\% & 40.7\% & 2.1\% & 76.4\% & 62.6\% & 80.3\% & - & 0.874 & 0.832 & 0.783 & - \\
REINVENT \cite{olivecrona2017molecular} & 99.3\% & 98.5\% & 97.4\% & 47.9\% & 61.0\% & 31.6\% & 39.7\% & 56.1\% & 0.733 & 0.729 & 0.595 & 0.621 \\
RationaleRL \cite{jin2020multi} & 100\% & 100\% & 100\% & 74.8\% & 53.4\% & 46.2\% & 97.3\% & 56.8\% & 0.888 & 0.862 & 0.824 & 0.701 \\ 
\midrule
\textbf{JANUS} & \textbf{100\%} & \textbf{100\%} & \textbf{100\%} & \textbf{100\%} & \textbf{82.1\%} & \textbf{40.9\%} & \textbf{77.8\%} & \textbf{32.6\%} & \textbf{0.885} & \textbf{0.894} & \textbf{0.875} & \textbf{0.821} \\ 
\textbf{JANUS+P} & \textbf{100\%} & \textbf{100\%} & \textbf{100\%} & \textbf{100\%} & \textbf{83.5\%} & \textbf{42.6\%} & \textbf{79.6\%} & \textbf{17.8\%} & \textbf{0.883} & \textbf{0.895} & \textbf{0.861} & \textbf{0.843} \\
\textbf{JANUS+C(50\%)} & \textbf{100\%} & \textbf{100\%} & \textbf{100\%} & \textbf{100\%} & \textbf{82.9\%} & \textbf{40.4\%} & \textbf{74.4\%} & \textbf{18.4\%} & \textbf{0.884} & \textbf{0.894} & \textbf{0.877} & \textbf{0.841} \\
\bottomrule
\end{tabular}}
\label{tab:imitated_inhibition}
\end{table}

Since these benchmarks provide a dataset to be used for training we initiated JANUS with the best molecules from this data, respectively. In addition, we augmented the fitness with a binary classifier that assigns a high fitness to molecules that fulfill all constraints. For this task, JANUS generates 5,000 unique molecules in each generation. Furthermore, we derived substructures from the reference molecules and use them as specific mutation rules in the genetic operators (cf. Figure \ref{fig:dataset_fragments}). The corresponding results are summarized in Table \ref{tab:imitated_inhibition}. JANUS achieves perfect, and hence state-of-the-art, performance for all four tasks in terms of success. While the diversity is also high in the generated molecules and comparable to literature baselines, novelty is significantly lower for some of the tasks. We believe that novelty can be further increased by running JANUS for more generations, by implementing a discriminator into JANUS \cite{nigam2019augmenting}, and by using the default mutation rules rather than deriving them from the reference data. Interestingly, we observe that many molecules that pass the constraints are very large. This suggests that there are problems in these benchmark tasks that classify large structures favorably. Some of the structures generated by JANUS in this benchmark are depicted in Figures \ref{fig:imitated_inhibition_structures_gsk3b} -- \ref{fig:imitated_inhibition_structures_gsk3b_jnk3_qed_sascore}.

\subsection{Molecular Docking}

\begin{figure}[htbp!]
\centering
\begin{tabular}{ ll }
(a) & (b) \\
\includegraphics[width=0.47\textwidth]{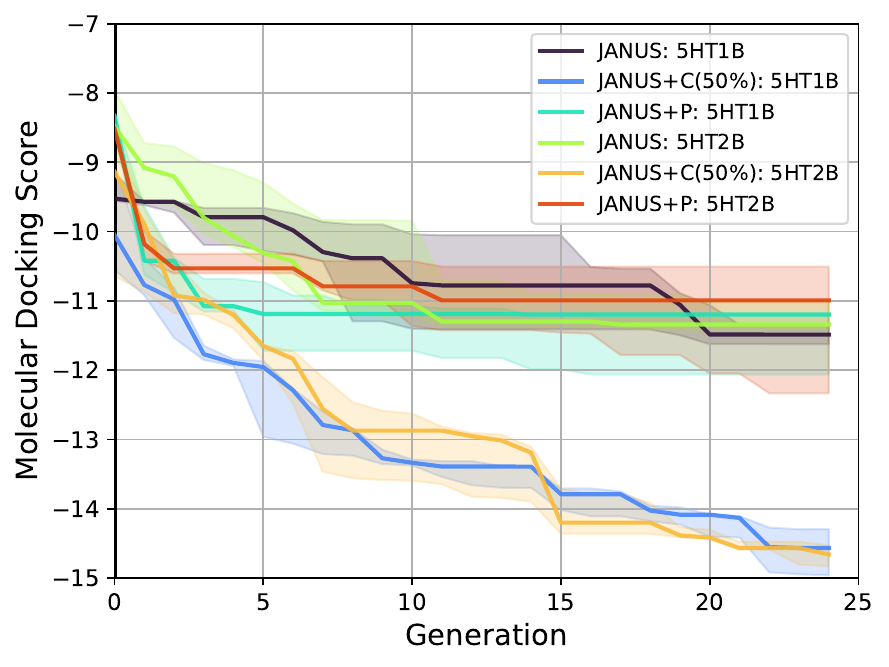} %
& %
\includegraphics[width=0.47\textwidth]{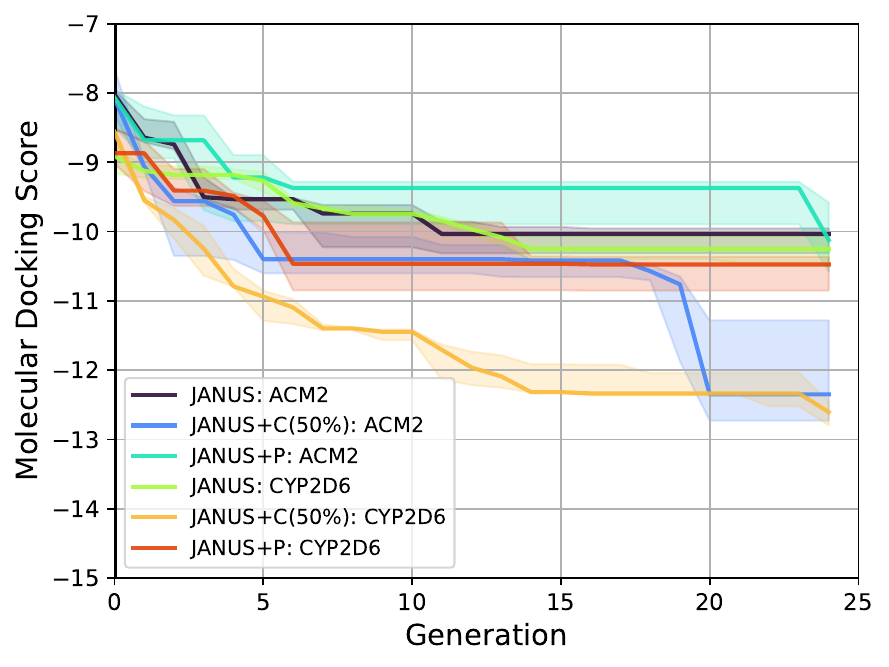}
\end{tabular}
\caption{Optimization progress of JANUS with three variations of selection pressure in the minimization of the docking scores to the protein targets (a) 5HT1B and 5HT2B, and (b) ACM2 and CYP2D6. Progress is depicted via the median of the highest fitness in each generation across 3 independent runs. The semi-transparent areas in both (a) and (b) depict the fitness intervals between the corresponding 10\% and 90\% quantiles.\label{fig:docking}}
\end{figure}

Better than trying to imitate molecular docking with a DNN classifier (that differentiates whether a molecule binds to a target protein or not), we conduct a more realistic experiment. Recently, to resemble the complexity of real-life drug discovery tasks, molecular docking has been proposed as a benchmarking task for generative models \cite{cieplinski2020we}. We used JANUS to design molecules minimizing the molecular docking score to the protein targets 5HT1B, 5HT2B, ACM2 and CYP2D6, respectively. The corresponding results in comparison to literature baselines are summarized in Table \ref{tab:docking}. Again, we run JANUS with three different variations of the selection pressure, first without any additional selection pressure (“JANUS”), second with a DNN predictor (“JANUS+P”) and third with a DNN classifier (“JANUS+C(50\%)”). The corresponding optimization progress is illustrated in Figure \ref{fig:docking}.

\begin{table}[htbp!]
\caption{Comparison of JANUS against literature baselines for the minimization of molecular docking scores to the protein targets 5HT1B, 5HT2B, ACM2 and CYP2D6. The first value corresponds to the docking score, the value in parenthesis is the diversity of the molecules generated. The entry denoted as “JANUS” does not use additional selection pressure for the exploration population, “JANUS+P” uses a DNN predictor as additional selection pressure, the “JANUS+C(50\%)” entry uses a DNN classifier as additional selection pressure.}
\resizebox{0.9\textwidth}{!}{
\begin{tabular}{ rcccc } 
\toprule
\textbf{Method} & \textbf{5HT1B} & \textbf{5HT2B} & \textbf{ACM2} & \textbf{CYP2D6} \\ 
\midrule
ZINC (10\%) &  -9.894 (0.862) &  -9.228 (0.851) & -8.282 (0.860) & -8.787 (0.853) \\
ZINC (1\%) &  -10.496 (0.861) &  -9.833 (0.838) & -8.802 (0.840) & -9.291 (0.894) \\
Train (10\%) &  -10.837 (0.749) &  -9.769 (0.831) & -8.976 (0.812) & -9.256 (0.869) \\
Train (1\%) &  -11.493 (0.849) &  -10.023 (0.746) & -10.003 (0.773) & -10.131 (0.763) \\
CVAE \cite{gomez2018automatic} &  -4.647 (0.907) &  -4.188 (0.913) & -4.836 (0.905) & - (-) \\
GVAE \cite{kusner2017grammar} &  -4.955 (0.901) &  -4.641 (0.887) & -5.422 (0.898) & -7.672 (0.714) \\
REINVENT \cite{olivecrona2017molecular} &  -9.774 (0.506) &  -8.657 (0.455) & -9.775 (0.467) & -8.759 (0.626) \\
\midrule
\textbf{JANUS}          & \textbf{-9.6 $\pm$ 0.9 (0.126)}     &    \textbf{-9.8 $\pm$ 0.7 (0.133)} & \textbf{-8.1 $\pm$ 0.5 (0.112)} &  \textbf{-9.1 $\pm$ 0.4 (0.166)} \\ 
\textbf{JANUS+P}        & \textbf{-9.9 $\pm$ 0.9 (0.132)}     &    \textbf{-9.8 $\pm$ 1.5 (0.166)} & \textbf{-8.0 $\pm$ 0.5 (0.125)} &  \textbf{-9.3 $\pm$ 0.6 (0.194)} \\  
\textbf{JANUS+C(50\%)} & \textbf{-13.8 $\pm$ 0.5 (0.366)}     &    \textbf{-13.8 $\pm$ 0.4 (0.331)} & \textbf{-9.9 $\pm$ 0.3 (0.235)} &  \textbf{-11.7 $\pm$ 0.4 (0.363)} \\  
\bottomrule
\end{tabular}}
\label{tab:docking}
\end{table}

The results summarized in Table \ref{tab:docking} show that JANUS readily proposes molecules with favorable docking scores for each of the protein targets. In particular, JANUS outperforms REINVENT significantly for all targets in terms of the docking scores, while only slightly for ACM2 when selection pressure is applied via a classifier, much more significantly for all other targets. However, while the molecular docking scores produced by JANUS are favourable, the diversities are significantly lower leaving room for further improvements. Notably, we limited these runs to 25 generations, which we believe contributes to the low observed diversities. Nevertheless, inspired by the GA+D approach by Nigam et al. \cite{nigam2019augmenting}, we believe that implementing a discriminator into JANUS would lead to significant improvements in diversity in the future. Furthermore, we observe that the runs with additional selection pressure from the DNN classifier have both significantly lower docking scores and higher diversities demonstrating its supremacy for both exploration and exploitation compared to the other selection pressure variations tested. Finally, we would like to point out that the molecules produced by JANUS contain several structural features that would be infeasible in a real drug such as extensive cummulene systems or triple bonds inside small rings. This is not a problem of JANUS but a deficiency of the benchmark task as no metric to assess stability or synthesizability is explicitly included. Hence, we propose to refine these molecular docking benchmarks by explicitly incorporating the SAscore in the final metric or by using a set of stability filters to assess the generated structures.

\section{Conclusion}
In this work, we presented JANUS, a genetic algorithm that propagates two populations, one explorative and one exploitative, and exchanges members between them. JANUS relies on the robustness of the SELFIES representation of molecular graphs and uses the STONED algorithm for efficient generation of molecules by string manipulations, requiring no domain knowledge. JANUS can apply additional selection pressure via an on-the-fly trained deep neural network that assesses the fitness of the generated molecules and propagating good members to subsequent generations, leading to faster optimization. Our model outperforms literature baselines in common inverse molecular design benchmarks that are relevant to drug discovery and material design. In particular, JANUS even outperforms previous computer-inspired human design in the maximization of the penalized logarithm of the octanol-water partition coefficient. 

Nevertheless, we see significant room for further improvement of JANUS. First, we plan to implement and refine the discriminator approach developed by Nigam et al. \cite{nigam2019augmenting} to avoid getting stuck in local optima and allow for more extensive exploration of the chemical space. Additionally, the current standard for multiobjective inverse molecular design is the (linear) combination of the objectives into one superobjective. However, this requires tailoring the weights of each objective to obtain best results. To avoid that, we will explore the implementation of more general and practical frameworks that concatenate multiple objectives into one superobjective out of the box such as Chimera \cite{hase2018chimera} or use alternative multi-objective GA approaches \cite{konak2006multi}.

\section{Acknowledgement}
R. P. acknowledges funding through a Postdoc.Mobility fellowship by the Swiss National Science Foundation (SNSF, Project No. 191127). A. A.-G. thanks Anders G. Fr{\o}seth for his generous support. A. A.-G. also acknowledges the generous support of Natural Resources Canada and the Canada 150 Research Chairs program. Computations were performed on the Cedar supercomputer situated at the Simon Fraser University in Burnaby. In addition, we acknowledge support provided by Compute Ontario and Compute Canada.

\medskip
\bibliographystyle{unsrt}
\bibliography{refs}

\newpage

\renewcommand{\thepage}{S\arabic{page}} 
\setcounter{page}{1}
\renewcommand{\thefigure}{S\arabic{figure}} 
\setcounter{figure}{0}
\renewcommand{\thetable}{S\arabic{table}} 
\setcounter{table}{0}
\renewcommand{\theequation}{S\arabic{equation}} 
\setcounter{equation}{0}
\renewcommand{\thesection}{S\arabic{section}} 
\setcounter{section}{0}

\section{Supplementary Information}

\subsection{Property Distributions of Random SELFIES}

\begin{figure}[htbp!]
\centering
\includegraphics[width=0.74\textwidth]{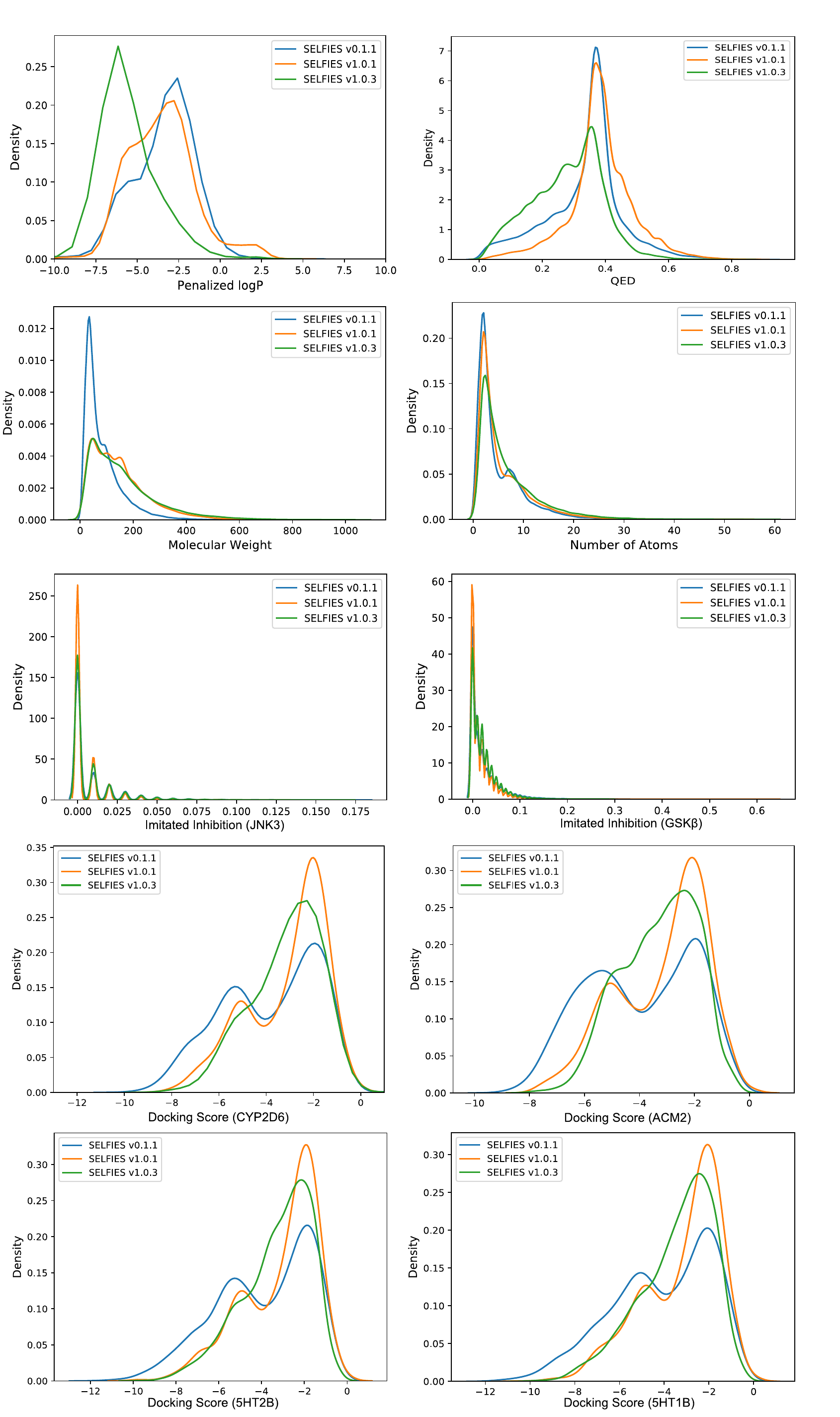}
\caption{Property distribution densities of molecules generated from random SELFIES using different versions of the SELFIES package.\label{fig:random_selfies}}
\end{figure}

The properties of molecules generated from random SELFIES depends on the specific version of the alphabet used. This does not only affect the properties of completely randomly generated structures but also the properties of randomly modified structures generated via point mutations of an initial structure. The dependence of the property distribution densities of various properties used in the course of this work on the version of SELFIES used is illustrated in Figure \ref{fig:random_selfies}.

\newpage

\subsection{Genetic Operations Using STONED}
\begin{figure}[htbp!]
\centering
\includegraphics[width=1.00\textwidth]{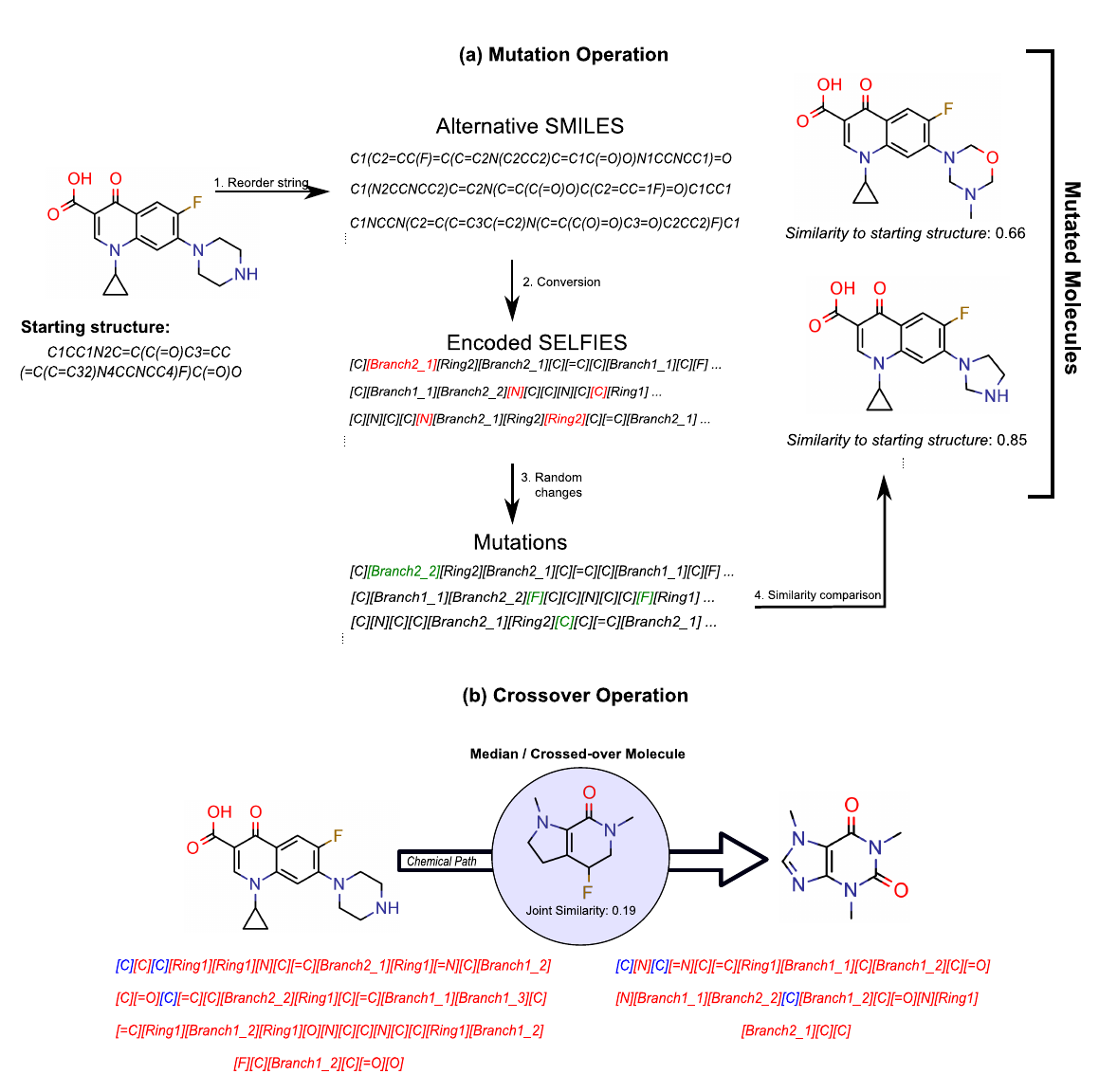}
\caption{Overview of the mutation and crossover operations implemented in JANUS. (a) A combination of SMILES reordering and random SELFIES modification yields a diverse set of mutated molecules. (b) Chemical paths between two SELFIES provide new child structures that combine elements of both parent molecules.\label{fig:stoned_scheme}}
\end{figure}

As discussed in the main text, JANUS uses the STONED algorithm for efficient structure generation based on the SELFIES representation of molecular graphs. The operations of STONED that we implemented into JANUS are illustrated in Figure \ref{fig:stoned_scheme}. Mutation of population members are performed via random string modifications of SELFIES. The generation of children from two parent molecules is carried out via the formation of chemical paths between them.

\newpage

\subsection{Extracting Fragments from Datasets}
\begin{figure}[htbp!]
\centering
\includegraphics[width=0.55\textwidth]{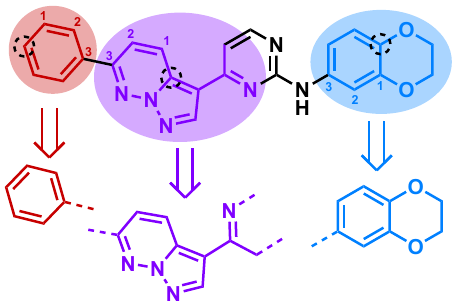}
\caption{Schematic illustration of the extraction of circular fragments from molecules with a radius of 3 from the circled atoms.\label{fig:dataset_fragments}}
\end{figure}

The automated extraction of fragments from datasets that is used as optional method to bias the genetic operators is illustrated in Figure \ref{fig:dataset_fragments}. The molecule depicted is taken from the training datasets of the imitated inhibition tasks.

\newpage

\subsection{Constrained Penalized log P Optimization}
\label{sec:constrained_logp}

For the constrained penalized log P optimization benchmark, we use the setup employed by Jin et al. \cite{jin2018junction}. The goal is to improve the penalized log P values of select molecules while maintaining similarity ($\delta$) constraints to them. This task has two subtask, one is to maintain a similarity of at least 0.4 to the original structure, the other is to maintain a similarity of at least 0.6. The tasks are assessed by the improvement in penalized log P values and by success, which is measured by the percentage of structures that were successfully improved in terms of penalized log P within the respective similarity constraints. In total, the benchmark task is to perform this constrained optimization for 800 molecules selected from the ZINC dataset. For each molecule, we run JANUS independently, and the generations are seeded by the starting structure. Each experiment is run for at most 10 generations. The corresponding results are summarized in Table \ref{tab:constr_plp}.

\begin{table}[htbp!]
\caption{Comparison on constrained improvement of penalized log P of specific molecules.}
\begin{tabular}{r cccc cccc}
\toprule
\textbf{Method} & \multicolumn{2}{c}{$\delta \geq$ 0.4} & \multicolumn{2}{c}{$\delta \geq$ 0.6} \\
       & \textbf{Improvement}       & \textbf{Success} & \textbf{Improvement}       & \textbf{Success}   \\
\midrule
JTVAE \cite{jin2018junction}            &  0.84 $\pm 1.45$  &   83.6\%  &   0.21 $\pm$ 0.71   & 46.4\% \\ 
GCPN  \cite{you2018graph}               & 2.49 $\pm$ 1.30   &   100.0\% & 0.79 $\pm$ 0.63  & 100.0\%   \\
MMPA \cite{jin2020multi}                &  3.29 $\pm$ 1.12  &  -  & 1.65 $\pm$ 1.44 &  -  \\
DEFactor \cite{assouel2018defactor}     & 3.41 $\pm$ 1.67   &   85.9\%   &  1.55 $\pm$ 1.19  &  72.6\%   \\
VJTNN \cite{jin2018learning}            &  3.55 $\pm$ 1.67  &   -       &  2.33 $\pm$ 1.17    &  - \\ 
GA \cite{nigam2019augmenting}           &  5.93 $\pm$ 1.41  &  100.0\%  &  3.44 $\pm$ 1.09 & 99.8\%\\     

\midrule
\textbf{JANUS} & \textbf{8.34 $\pm$ 3.17} & \textbf{100\%} & \textbf{5.29 $\pm$ 2.33} & \textbf{100\%}  \\ 
\bottomrule
\end{tabular}
\label{tab:constr_plp}
\end{table}

\begin{figure}[htbp!]
\centering
\includegraphics[width=1.0\textwidth]{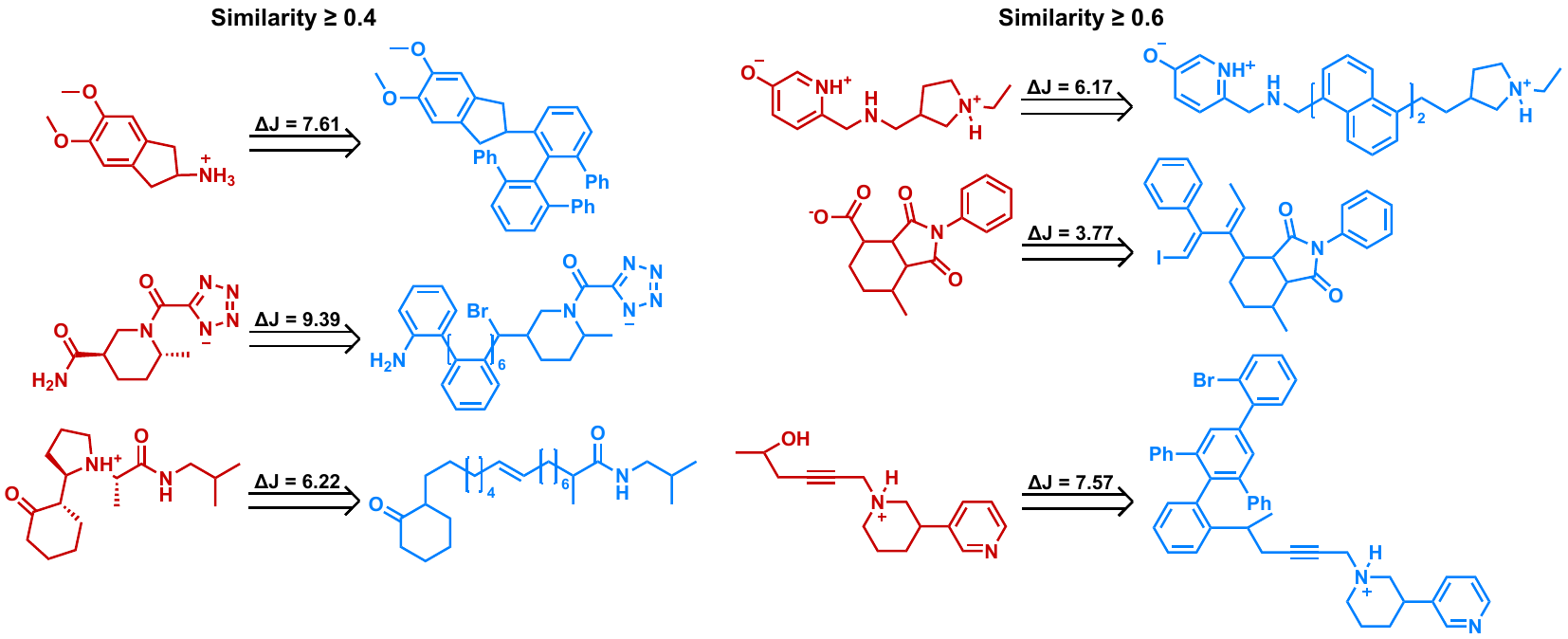}
\caption{Selection of structures from the constrained penalized log P (J) optimization task. Red structures are the starting structures, blue structures are the best structures generated by JANUS.\label{fig:constrained_logp}}
\end{figure}

It should be noted that JANUS achieved state-of-the-art performance in this benchmark with a lower number of generations than the GA approach developed by Nigam et al. \cite{nigam2019augmenting} which was run for 20 generations. Particular pairs of initial structures and structures with maximized constrained penalized log P generated by JANUS are depicted in Figure \ref{fig:constrained_logp}. They show that the GA finds a few very effective modifications that improve the penalized log P while maintaining a sufficient similarity. In particular, keeping less polar substructures but replacing more polar parts with extensive aliphatic or aromatic moieties is particularly effective in the examples provided.

\newpage

\subsection{Imitated Inhibition}
\begin{figure}[htbp!]
\centering
\includegraphics[scale=0.7]{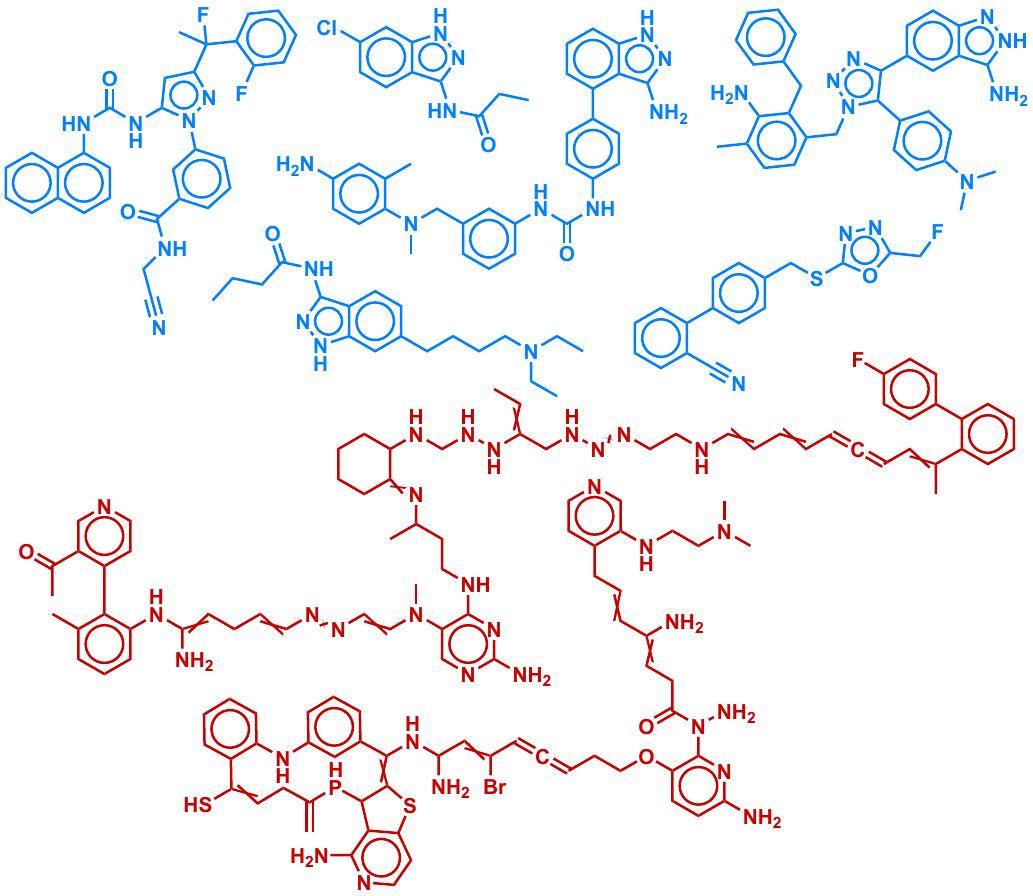}
\caption{Selection of structures from the imitated inhibition task optimizing for GSK3{\textbeta} inhibition. Blue structures are considered reasonable, red structures are considered problematic in terms of stability.\label{fig:imitated_inhibition_structures_gsk3b}}
\end{figure}

\begin{figure}[htbp!]
\centering
\includegraphics[scale=0.7]{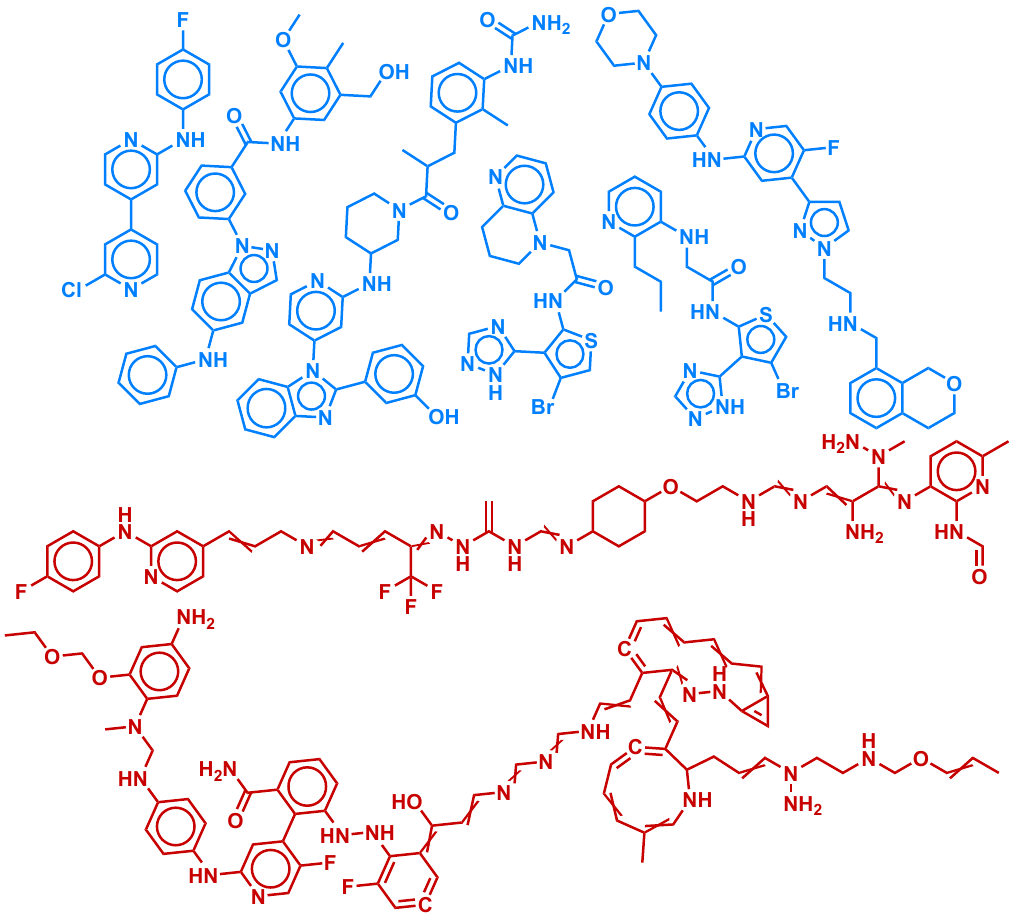}
\caption{Selection of structures from the imitated inhibition task optimizing for JNK3 inhibition. Blue structures are considered reasonable, red structures are considered problematic in terms of stability.\label{fig:imitated_inhibition_structures_jnk3}}
\end{figure}

\begin{figure}[htbp!]
\centering
\includegraphics[scale=0.7]{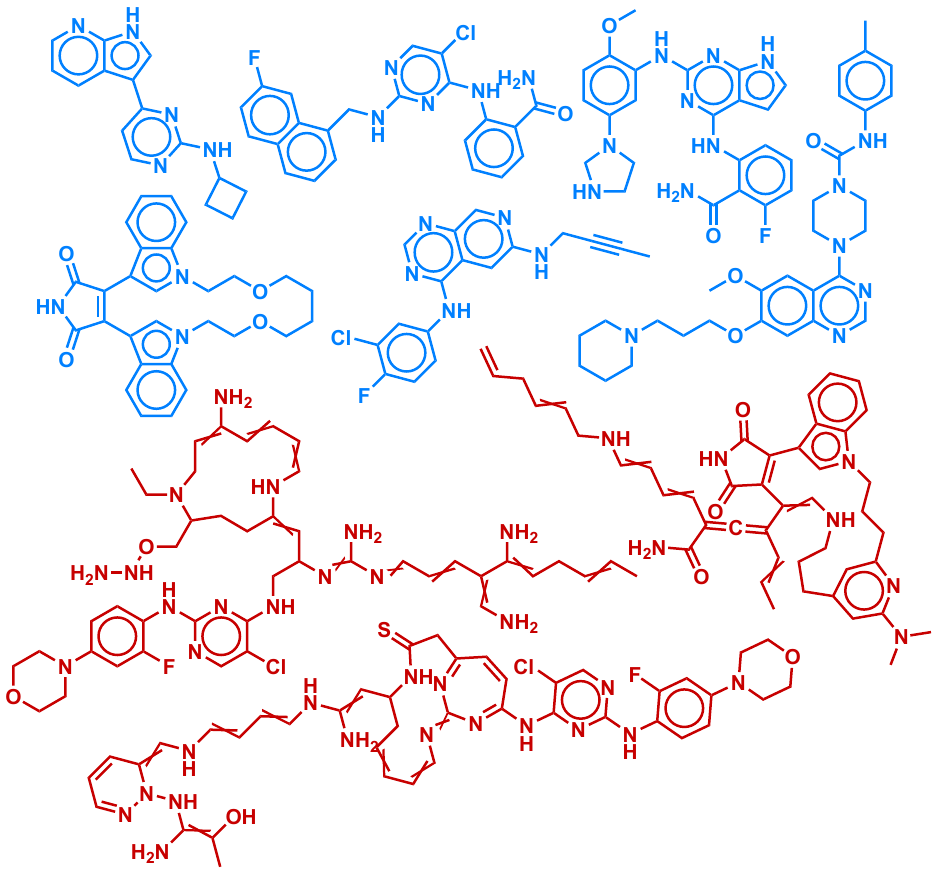}
\caption{Selection of structures from the imitated inhibition task optimizing for both GSK3{\textbeta} and JNK3 inhibition. Blue structures are considered reasonable, red structures are considered problematic in terms of stability.\label{fig:imitated_inhibition_structures_gsk3b_jnk3}}
\end{figure}

\begin{figure}[htbp!]
\centering
\includegraphics[scale=0.7]{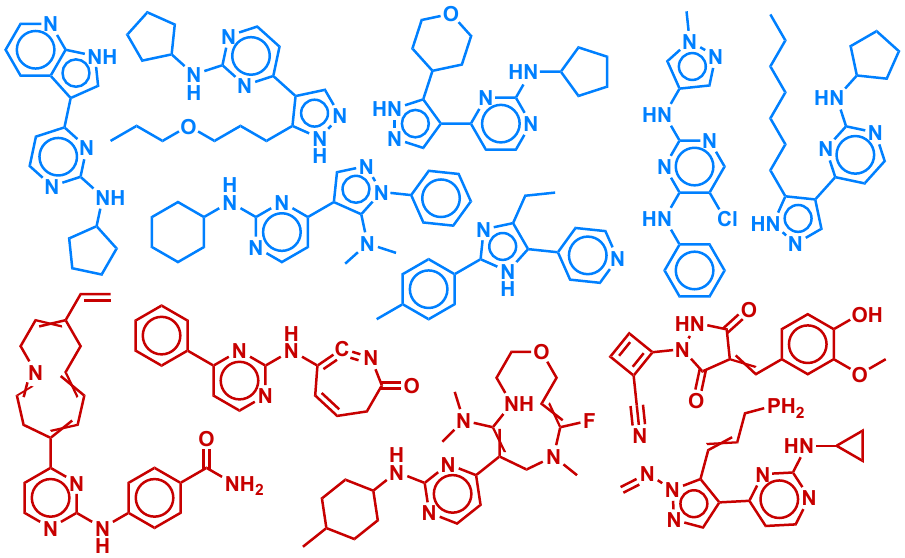}
\caption{Selection of structures from the imitated inhibition task optimizing for both GSK3{\textbeta} and JNK3 inhibition as well as QED and the SAscore. Blue structures are considered reasonable, red structures are considered problematic in terms of stability.\label{fig:imitated_inhibition_structures_gsk3b_jnk3_qed_sascore}}
\end{figure}

For this task, JANUS is initiated by molecules classified as inhibitor that are provided as a reference dataset in the benchmark. Across multiple generations, a list of molecules that fulfill all the criteria within each experiment is maintained. The mutation and crossover operations are carried out using one random structure selected from this list to replace a molecule from the population. Diversity and novelty of the 5,000 molecules generated are calculated based on the following expressions:
\begin{equation}
\text{Diversity} = 1 - \frac{2}{n(n-1)} \sum_{X,Y} \text{sim}(X,Y)\;
\end{equation}

\begin{equation}
\text{Novelty} =  \frac{1}{n}\sum_{G} 1 \left[\text{sim}(G, G_{\text{SNN}}) < 0.4\right]\;
\end{equation}
The expression $\text{sim}(X,Y)$ computes the pairwise molecular similarity for all $n$ structures calculated as the Tanimoto distance of the Morgan fingerprint (calculated with a radius of size 3 and a 2048 bit size). In addition, for a given compound $G$ fulfilling the requirements of the benchmark, $G_{\text{SNN}}$ refers to the molecule from the reference dataset that is closest to $G$ in terms of molecular similarity. Figures \ref{fig:imitated_inhibition_structures_gsk3b} -- \ref{fig:imitated_inhibition_structures_gsk3b_jnk3_qed_sascore} illustrate some structures produced by JANUS in the imitated inhibition tasks. Blue structures are considered reasonable, red structures are considered problematic based on expert opinion. It should be noted that some very large  and unstable structures are considered favorable inhibitors.

\newpage

\newpage

\newpage

\subsection{GuacaMol}
For the GuacaMol benchmark suite \cite{brown2019guacamol}, we set the generation size to 10,000 and initiated the population with the top 10,000 molecules from the provided reference dataset. To reduce the computational requirements, we only used mutations as genetic operations. Moreover, no additional selection pressure via a neural network was applied. The local search within the exploitative population is only performed on the fittest member of the explorative population. The results of JANUS on all the benchmark tasks is summarized in Table \ref{tab:guacamol_summary} and compared to various generative models from the literature.

\begin{table}[htbp!]
\caption{Comparison of JANUS against literature baselines for the GuacaMol benchmark suite \cite{brown2019guacamol}. The entry denoted as “JANUS” does not use additional selection pressure for the exploration population. The following abbreviations are used: redisc. for rediscovery, sim. for similarity.}
\resizebox{1.\textwidth}{!}{
\begin{tabular}{lcccccccccc}
\toprule
 & \textbf{STONED} & \textbf{SMILES} & \textbf{SMILES} & \textbf{CReM} & \textbf{Graph} & \textbf{MSO} & \textbf{EvoMol} & \textbf{MolFinder} & \textbf{GEGL} & \textbf{JANUS}\\
\textbf{Benchmark} & \cite{nigam2021beyond} & \textbf{GA} \cite{brown2019guacamol} & \textbf{LSTM} \cite{brown2019guacamol} & \cite{polishchuk2020crem} & \textbf{GA} \cite{brown2019guacamol} & \cite{winter2019efficient} & \cite{leguy2020evomol} & \cite{kwon2021molfinder} & \cite{NEURIPS2020_8ba6c657}  & (here) \\
\midrule
Celecoxib redisc.      &   0.556 &  0.732 &  1.000 &  1.000 &  1.000 &  1.000 &  1.000 &  1.000 &  1.000 &  1.000 \\
Troglitazone redisc.   &   0.543 &  0.515 &  1.000 &  1.000 &  1.000 &  1.000 &  1.000 &  1.000 &  1.000 &  1.000 \\
Thiothixene redisc.    &   0.677 &  0.598 &  1.000 &  1.000 &  1.000 &  1.000 &  1.000 &  1.000 &  1.000 &  1.000 \\
Aripiprazole sim.    &   0.716 &  0.834 &  1.000 &  1.000 &  1.000 &  1.000 &  1.000 &  1.000 &  1.000 &  1.000 \\
Albuterol sim.       &   0.939 &  0.907 &  1.000 &  1.000 &  1.000 &  1.000 &  1.000 &  1.000 &  1.000 &  1.000 \\
Mestranol sim.       &   0.769 &  0.790 &  1.000 &  1.000 &  1.000 &  1.000 &  1.000 &  1.000 &  1.000 &  1.000 \\
C\textsubscript{11}H\textsubscript{24}                     &   1.000 &  0.829 &  0.993 &  0.966 &  0.971 &  0.997 &  1.000 &  1.000 &  1.000 &  1.000 \\
C\textsubscript{9}H\textsubscript{10}N\textsubscript{2}O\textsubscript{2}PF\textsubscript{2}Cl             &   0.886 &  0.889 &  0.879 &  0.940 &  0.982 &  1.000 &  1.000 &  1.000 &  1.000 &  1.000 \\
Median molecules 1         &   0.351 &  0.334 &  0.438 &  0.371 &  0.406 & 0.437 & 0.455 & 0.412 & 0.455 &  0.434 \\
Median molecules 2         &   0.395 &  0.380 &  0.422 &  0.434 &  0.432 & 0.395 & 0.417 & 0.454 & 0.437 &  0.416 \\
Osimertinib MPO            &   0.863 &  0.886 &  0.907 &  0.995 &  0.953 & 0.966 & 0.969 & 0.945 & 1.000 & 0.947 \\
Fexofenadine MPO           &   0.878 &  0.931 &  0.959 &  1.000 &  0.998 & 1.000 & 1.000 & 0.999 & 1.000 & 0.999 \\
Ranolazine MPO             &   0.812 &  0.881 &  0.855 &  0.969 &  0.920 & 0.931 & 0.957 & 0.947 & 0.958 & 0.920 \\
Perindopril MPO            &   0.629 &  0.661 &  0.808 &  0.815 &  0.792 & 0.834 & 0.827 & 0.816 & 0.882 & 0.817 \\
Amlodipine MPO             &   0.738 &  0.722 &  0.894 &  0.902 &  0.894 & 0.900 & 0.869 & 0.924 & 0.924 & 0.905 \\
Sitagliptin MPO            &   0.592 &  0.689 &  0.545 &  0.763 &  0.891 & 0.868 & 0.926 & 0.948 & 0.922 & 0.901 \\
Zaleplon MPO               &   0.674 &  0.413 &  0.669 &  0.770 &  0.754 & 0.764 & 0.793 & 0.695 & 0.834 & 0.774 \\
Valsartan SMARTS           &   0.864 &  0.552 &  0.978 &  0.994 &  0.990 & 0.994 & 0.998 & 0.999 & 1.000 & 0.993 \\
Deco hop                   &   0.968 &  0.970 &  0.996 &  1.000 &  1.000 &  1.000 &  1.000 &  1.000 &  1.000 &  1.000 \\
Scafold hop                &   0.854 &  0.885 &  0.998 &  1.000 &  1.000 &  1.000 &  1.000 &  0.948 &  1.000 &  1.000 \\
\midrule
\textbf{Total Score}                &  14.704 & 14.396 & 17.340 & 17.919 &  17.983 & 18.086 & 18.211 & 18.087 & 18.412 & 18.106\\
\bottomrule
\end{tabular}}
\label{tab:guacamol_summary}
\end{table}

\end{document}